# Matrix Profile XXII: Exact Discovery of Time Series Motifs under DTW

Sara Alaee
University of California
Riverside
salae001@ucr.edu

Kaveh Kamgar
University of California
Riverside
kkamg001@ucr.edu

Eamonn Keogh
University of California
Riverside
eamonn@cs.ucr.edu

*Abstract*—Over the last decade, time series motif discovery has emerged as a useful primitive for many downstream analytical tasks, including clustering, classification, rule discovery, segmentation, and summarization. In parallel, there has been an increased understanding that Dynamic Time Warping (DTW) is the best time series similarity measure in a host of settings. Surprisingly however, there has been virtually no work on using DTW to discover motifs. The most obvious explanation of this is the fact that both motif discovery and the use of DTW can be computationally challenging, and the current best mechanisms to address their lethargy are mutually incompatible. In this work, we present the first scalable exact method to discover time series motifs under DTW. Our method automatically performs the best trade-off between time-to-compute and tightness-of-lower-bounds for a novel hierarchy of lower bounds representation we introduce. We show that under realistic settings, our algorithm can admissibly prune up to 99.99% of the DTW computations.

## I. INTRODUCTION

Time series motif discovery — the unearthing of locally conserved behavior in a long time series — has emerged as one of the most important time series primitives in the last decade [1]. In recent years, there has been significant progress in the scalability of motif discovery, but essentially all algorithms use the Euclidean Distance (ED) [2][6]. This is somewhat surprising, because in parallel, the community seems to have converged on the understanding that the Dynamic Time Warping (DTW) is superior in most domains, at least for the tasks of clustering, classification, and similarity search [3][8][9][10]. Could DTW also be superior to ED for motif discovery? To preview our answer to this question, consider Fig. 1, which shows the top-1 motif discovered in an electrical power demand dataset, using the Euclidean distance [7].

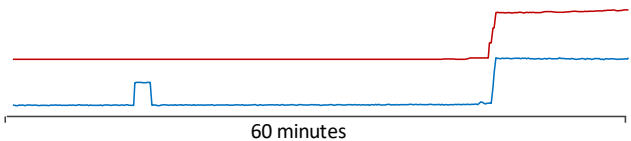

Fig. 1. The top-1 Euclidean distance motif discovered in a one-month long electrical power demand dataset. The full dataset that this motif was extracted from, like all other datasets used in this paper, is available at [16].

We have no obvious reasons to discount this motif. It clearly shows the highly conserved behavior. However, now let us consider Fig. 1, which shows a different pair of subsequences from the same dataset. In retrospect, we would surely have preferred to have discovered this pair of motifs as the top-1 motif. The complexity of the pattern that is conserved points to a common mechanism. In fact, this *is* the case. This pattern corresponds to a particular program from a dishwasher. Why was this pattern not discovered by the classic motif discovery algorithm? As we will show, the use of ED is the culprit and DTW is the solution.

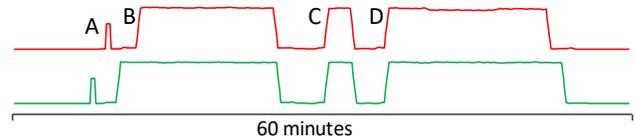

Fig. 2. A pair of subsequences from household electrical power demand data. The pattern corresponds to a particular dishwasher cycle: (A) short run of discharge pump to empty any liquid in the machine, (B) pumping water into reservoir, (C) spraying water over dishes (D) pumping out water.

As we will show, given the ability to find motifs under DTW, examples like the one above are replete in diverse domains such as industry, medicine, and human behavior. Given that there is a large body of literature on both motif discovery and DTW, why are there essentially no DTW-based motif discovery tools?

We believe that the following explains this omission. Both motif discovery and DTW comparisons are famously computationally demanding [1][3]. Recent years have seen significant progress for both, especially the *Matrix Profile* for the former [13], but the main speed-up techniques for each are not obviously combinable.

In this work we introduce a novel algorithm that makes DTW motif discovery tenable for large datasets for the first time. We call our algorithm SWAMP, Scalable Warping Aware Matrix Profile. This is something of a misnomer, since we attempt to *avoid* computing most of the true DTW Matrix Profile by instead computing much cheaper upper/lower bounding Matrix Profiles. We claim the following contributions:

- We show, for the first time, that there exists conserved structure in real-world time series that can be found with DTW motifs, but not with classic Euclidean distance motifs [6]. It was not clear that this had to be the case, as [6] and others had argued for the diminished utility of DTW for *motif discovery* (*all*-to-all search), relative to its known utility for similarity search (*one*-to-all search).

- We introduce SWAMP, the first *exact* algorithm for DTW motif discovery that significantly outperforms brute force search by two or more orders of magnitude.

The rest of the paper is organized as follows. In Section II and Section III, we present the formal definitions and background, before outlining our approach in IV. Section V contains experimental evaluations. Finally, we offer conclusions and directions for future work in Section VI.

## II. DEFINITIONS AND NOTATIONS

We begin by introducing the necessary definitions and fundamental concepts, beginning with the definition of a *Time Series*:

**Definition 1**: A Time Series $T = t_1, t_2, \ldots, t_n$ is a sequence of $n$ real values.

Our distance measures quantify the distance between two time series based on local subsections called *subsequences*:

**Definition 2**: A subsequence $T_{i,L}$ is a contiguous subset of values with length $L$ starting from position $i$ in time series T; the subsequence $T_{i,L}$ is in form $T_{i,L} = t_i, t_{i+1}, \ldots, t_{i+L-1}$, where $1 \leq i \leq n - L + 1$ and $L$ is a user-defined subsequence length with value in range of $4 \leq L \leq |T|$.

Here we allow $L$ to be as short as four, although that value is pathologically short for almost any domain [8].

The nearest neighbor of a subsequence is the subsequence that has the smallest distance to it. The closest pairs of these neighbors are called the time series *motifs*.

**Definition 3**: A motif is the most similar subsequence pair of a time series. Formally, $T_{a,L}$ and $T_{b,L}$ is the motif pair iff $dist(T_{a,L}, T_{b,L}) \leq dist(T_{i,L}, T_{j,L}) \forall i, j \in [1,2,\ldots,n - L + 1]$, where $a \neq b$ and $i \neq j$, and $dist$ is a distance measure.

One can observe that the potential best matches to a subsequence (other than itself) tend to be the subsequences beginning immediately before or after the subsequence. However, we clearly want to exclude such redundant "near self matches". Intuitively, any definition of motif should exclude the possibility of counting such *trivial matches*.

**Definition 4**: Given a time series T, containing a subsequence $T_{i,L}$ beginning at position $i$ and a subsequence $T_{j,L}$ beginning at $j$, we say that $T_{j,L}$ is a trivial match to $T_{i,L}$ if $j \leq i + L - 1$.

Following [2] we use a vector called the *Matrix Profile* (*MP*) to represent the distances between all subsequences and their nearest neighbors.

**Definition 5**: A Matrix Profile (MP) of time series T is a vector of distances between each subsequence $T_{i,L}$ and its nearest neighbor (closest match) in time series T.

The classic Matrix Profile definition assumes Euclidean distance measure which computes the distance between the i$^{th}$ point in one subsequence with the i$^{th}$ point in the other (see Fig. 3.*left*). However, as shown in Fig. 3.*center*, the non-linear DTW alignment allows a more intuitive distance that matches similar shapes even if they are locally out of phase. For brevity, we omit a formal definition of the (increasingly well-known) DTW, instead referring the interested reader to [9][3][8].

Similarity search under DTW can be demanding in terms of CPU time. One way to address this problem is to use a *lower bound* to help prune sequences that could not possibly be a best match [8]. While there exist dozens of lower bounds in the literature, in our work we use a generalization of the $LB_{Keogh}$ [3][8].

**Definition 6**: The $LB_{Keogh}$ lower bound between a time series Q and another time series T, given a warping window size $w$, is defined as the distance from the closest of the upper and lower envelopes around Q, to T, as in (1):

$$LB_{Keogh}(Q,T) = \sqrt{\sum_{i=1}^{n} \begin{cases} (t_i - U_i)^2 & if\ t_i > U_i \\ (t_i - L_i)^2 & if\ t_i < L_i \\ 0 & otherwise \end{cases}} \quad (1)$$

Where the upper envelope ($U_i$) and lower envelope ($L_i$) of Q are defined as in (2):

$$\begin{aligned} U_i &= max(q_{i-w}, q_{i-w+1}, \ldots, q_{i+w}) \\ L_i &= min(q_{i-w}, q_{i-w+1}, \ldots, q_{i+w}) \end{aligned} \quad (2)$$

Fig. 3 illustrates this definition.

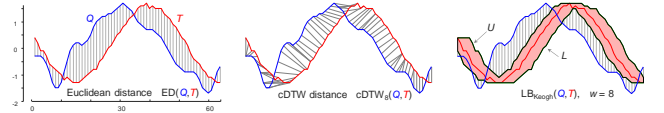

Fig. 3. For two time series Q and T: *left*) Their Euclidean Distance. *center*) Their DTW distance. *right*) Their $LB_{Keogh}$ distance.

For computationally demanding tasks, even the lower bound computation may take a lot of time. Thus, we plan to exploit a "spectrum" of lower bounds as we explain in Section IV.A, each of which makes a different compromise of fidelity versus tightness (defined in (4)).

To create this spectrum, we exploit our ability to perform various computations on the reduced dimensionality data. More concretely, we can perform downsampling using the *Piecewise Aggregate Approximation* (*PAA*) [4][11].

**Definition 7**: The PAA of time series T of length $n$ can be calculated by dividing T into $k$ equal-sized windows and computing the mean value of data within each window.

It is convenient to express the compression rate of a PAA approximation as "$D$ to 1", or $D:1$, where $D = \frac{n}{k}$. This notation can be visualized as shown in Fig. 4.

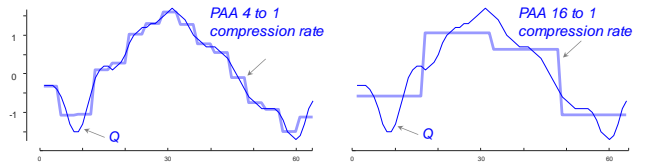

Fig. 4. A time series Q, downsampled using PAA to two different compression rates. *left*) 4:1 *right*) 16:1.

Given that we can downsample time series [4], we can also generalize $LB_{Keogh}$ to such downsampled data, with $LB_{Keogh}D{:}1$ ($D \geq 1$):

**Definition 8**: The downsampled lowerbound $LB_{Keogh}D:1(Q, T)$ between a time series Q and another time series T is defined as the distance from the closest of the downsampled upper and lower envelopes around Q, to the downsampled T. Formally as in (3):

$$LB_{Keogh}D:1(Q,T) = \sqrt{\sum_{i=1}^{n} \begin{cases} (t_{D_i} - U_{D_i})^2 & if\ t_{D_i} > U_{D_i} \\ (t_{D_i} - L_{D_i})^2 & if\ t_{D_i} < L_{D_i} \\ 0 & otherwise \end{cases}} \quad (3)$$

$T_D = PAA(T, D), U_D = PAA(U_Q, D), and\ L_D = PAA(L_Q, D)$

Fig. 5 illustrates this definition.

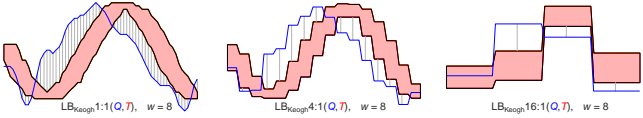

Fig. 5. An illustration of parametrized $LB_{Keogh}$. Three possible settings that make different trade-offs on the spectrum of time-to-compute vs. tightness of lower bound. The special case of $LB_{Keogh}1:1$ is the classic lower bound also shown in Fig. 3, and used extensively in the community [3][8][9].

Given these downsampled lower bounds, we can still use the $LB_{Keogh}$ distance, but we need to scale the distance by $\frac{\sqrt{n}}{D}$ to generate a tighter, yet still admissible lower bound. The proof of this variation of the lower bound appears in a slightly different context in [15]. To see why it is needed, refer to Fig. 5.*right*. Here each gray hatch-line represents the aggregate distance for 16 datapoints. If we only counted each line once, we would have a very weak lower bound. It *seems* that we could scale each line's contribution by 16 (or more generally, $D$), but then we would not have an admissible bound. It can be shown that $\frac{\sqrt{n}}{D}$ is the optimally tight *admissible* scaling [15].

To see how the parameterization affects the tightness of the lower bound, we selected 256 random pairs from the electrical demand dataset and computed both their true distance and the lower bound distances at the dimensionalities shown in Fig. 5. The results are shown in Fig. 6.

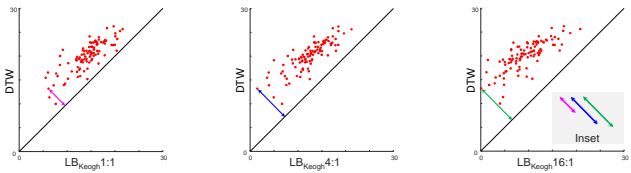

Fig. 6. An illustration of the tightness of the parametrized $LB_{Keogh}$. The tightness for each pair is inversely proportional to orthogonal distance to the diagonal line. For one randomly selected point, we show how this changes (inset), with longer lines indicating looser lower bounds.

Note that while our examples use powers of two for both the original and reduced dimensionality, PAA and our parametrized lower bounds are defined in the more general case [11].

### III. RELATED WORK

There is a huge body of literature on DTW [8] and on motif discovery [1]. However, there are very few papers on the intersection of these ideas.

Lagun et. al. created an algorithm to explore cursor movement data [5]. The algorithm discovers "common motifs" and does use the DTW distance, but it is better seen as a clustering algorithm that produces centroids that could be considered motifs. These papers speak to the utility of both motif discovery and to the use of DTW. However, these works do not offer us actionable insights for the task-at-hand.

### IV. OBSERVATIONS AND ALGORITHMS

Before introducing our algorithm in detail, we will take the time to outline the intuition behind our approach. In Fig. 7 we show a time series and its DTW MP.

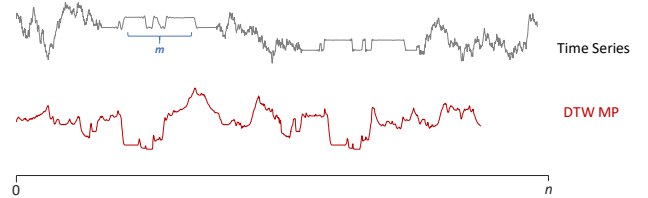

Fig. 7. A time series and its DTW MP. The lowest points of the DTW MP are the locations of the top-1 DTW motifs.

The two lowest points (they must have tied values by definition [13]) correspond to the top-1 DTW motif. Thus, while we have solved our task-at-hand, this brute force computation of the DTW MP required an untenable $O(n^2m^2)$ time.

There are some optimizations (which we use) including early abandoning, using the squared distance, etc. (see [8] and [7]). However, these only shave off small constant factors. Note that the Euclidean distance is an *upper* bound for the DTW. Moreover, there are perhaps a few dozen known *lower* bounds to DTW, including the $LB_{Keogh}$ [3]. In Fig. 8 we revisit the data shown in Fig. 7 to include the MPs for these two additional measures. Note that they "squeeze" the DTW MP.

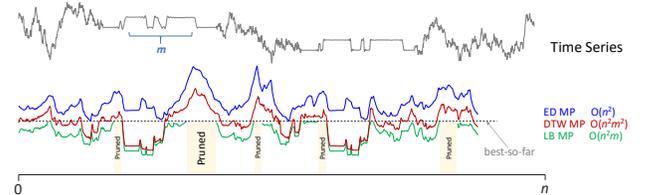

Fig. 8. A time series and its ED MP, its DTW MP and its $LB_{Keogh}1:1$MP. Note that the lowest values in the ED MP (denoted with the horizontal dashed line) are an upper bound on the values of the top DTW motif.

This figure suggests an immediate improvement to the brute force algorithm. The lowest value of the ED MP is an upper bound on the value of the top-1 DTW motif. Thus, before we compute the DTW MP, we could first compute the ED MP and use its smallest value to initialize the *best-so-far* value for the DTW MP search algorithm. This has two exploitable consequences. It would speed up the brute force algorithm, because the effectiveness of early abandoning is improved if you can find a good *best-so-far* early on. However, there is a much more consequential observation. Any region in the time series for which the lower bound is greater than the *best-so-far* can be admissibly pruned from the search space.

Note that this pruning can dramatically accelerate our search. For example, suppose that the fraction $p$ of the time series is pruned from consideration as the location of the best motif. We then only have to compute $(1-p)^2$ of the possible pairs of subsequences. Moreover, this ratio can only get better, as we find good matches that further drive the *best-so-far* down.

### A. Creating a Spectrum of Lower Bounds

As noted above, our SWAMP algorithm depends on the availability of multiple lower bounds that make different tradeoffs on the spectrum of tightness versus speed of execution. It is not meaningful to measure the tightness of lower bounds on a single pair of time series, as the idiosyncrasies of the particular pair of subsequences may favor different lower bounds. Instead, it is common to measure the tightness of a lower bound by averaging over many pairs of randomly chosen time series [9].

$$tightness(A,B) = \frac{LB(A,B)}{DTW(A,B)} \quad (4)$$

In Fig. 9 we average over six million pairs of random-walk for each setting.

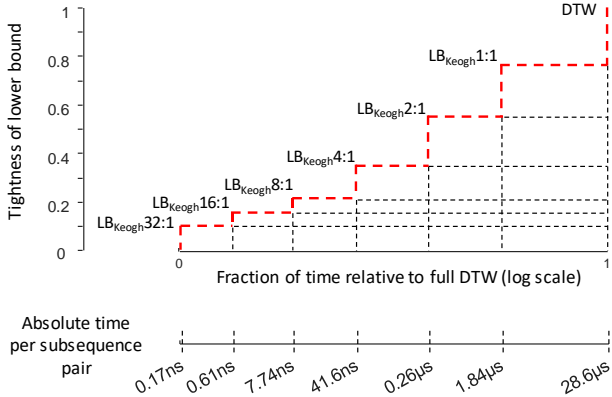

Fig. 9. A spectrum of lower bounds for DTW plotted with time (note the log scale) versus tightness. Recall that DTW is a lower bound to itself, thus occupies the top right corner.

It is important to ward off a possible misunderstanding. If we computed the *entire* $LB_{Keogh}1:1$ and found that it aggressively pruned off all but one pair of subsequences (the true top-1 motif), then we would achieve a speedup of about 28.6μs/1.84μs = 15.5. This 15-fold speed would be impressive, but it *appears* to be the upper bound on speed-up. However, as hinted at above, we hope to prune off many of the $LB_{Keogh}1:1$ computations themselves, with the much cheaper $LB_{Keogh}2:1$ calculations. Moreover, we plan to do this iteratively, using cheaper (but weaker) lower bounds to prune off as many as possible more expensive (but stronger) lower bounds.

### B. Introducing SWAMP

For notational simplicity we consider the task of finding the top-1 motif under DTW for a given value of $w$. The generalizations to top-K motifs or range motifs are trivial [13].

We can best think of SWAMP as a two-phase algorithm. In phase I it uses a single *upper* bound, and an adaptive hierarchy of *lower* bounds to prune off as many of the candidate time series subsequences as possible (*candidate* for being one of the best DTW motif pairs). Then, in phase II, any surviving pairs of subsequences are searched with a highly optimized "brute force" search algorithm. The algorithm in TABLE III formalizes SWAMP which includes subroutines that compute phase I and phase II.

### C. Phase I

We start by reviewing the two exploitable facts that we previewed in Fig. 8.

- The ED MP is the *upper bound* for $LB_{Keogh}$MP.
- The $LB_{Keogh}$MP is the *lower bound* for DTW MP.

Based on these observations, we know that any section of $LB_{Keogh}$MP (i.e. $LB_{Keogh}1:1$MP) that is greater than the minimum of ED MP (we consider that as the *best-so-far*), could not contain the best motif and can therefore be pruned. We can compute the DTW score for the region suggested by the lowest value of the pruned $LB_{Keogh}1:1$MP. If the score is lower than the minimum of ED MP, we can further lower the *best-so-far*. In this case, we can further reduce the number of DTW tests.

This basic strategy gains speedup, replacing most of the expensive DTW calculations with cheaper lower bound calculations. However, while computing $LB_{Keogh}1:1$MP is much faster than full DTW, it is still computationally expensive. Nevertheless, as we discussed in the previous section, we may not need to compute the full $LB_{Keogh}1:1$MP to find the best motifs. Instead, we can apply the above strategy on a hierarchy of cheaper downsampled $LB_{Keogh}$MP. The algorithm in TABLE I formalizes this process.

TABLE I. COMPUTEDSMP: HIERARCHICALLY COMPUTES THE DOWNSAMPLED LOWER BOUND MATRIX PROFILE AND PRUNES THE UNPROMISING LOCATIONS.

```
Algorithm: ComputeDSMP
Inputs: time series T, subsequence length L, warping window w
Outputs: the expanded LB_Keogh D:1 and indexes LBMP, LB_index,
the pruned locations of time series semantic matrix index
pruned, the candidate motif distance best-so-far
1    ED_mp ← ComputeMatrixProfile(T,L)  // Using SCRIMP
2    ED_motif_idx ← argmin(ED_mp)
3    best-so-far ← dtw_distance(ED_motif_idx)
4    D ← L
5    pruned(:) ← false
6    while D>0:   // iterate over increasing fine approximations
7       [LBMP,LB_index] ← LB_KeoghDSMP(T,L,D,pruned,w) // See TABLE II
8       LB_motif_dist, LB_motif_idx ← min(LBMP)
9       if LB_motif_dist < best-so-far:
10          best-so-far ← LB_motif_dist
11         pruned(LBMP > best-so-far) ← true
12      D ← floor(D/2)   // next iteration will be twice as fine
13   end
```

In line 1 we compute the classic Matrix Profile for the time series T with the given subsequence length $L$. This is needed to provide the upper bound of the distance between the DTW motifs we will discover. Using this Matrix Profile, we find the ED motifs, i.e. the pair of lowest values [13][14]. We then measure the distance between those motifs using the DTW distance rather than the ED distance, in order to initialize the *best-so-far* distance (lines 2-3).

Starting with a downsampling factor equal to the subsequence length (line 4), we first compute a very cheap lower bound for the entire time series using the algorithm in TABLE

II. If the DTW distance for the region suggested by the lowest value of this lower bound is smaller than the *best-so-far*, we update the *best-so-far*. For regions where it is too weak to prune, we selectively compute a tighter bound and repeat the same process. The algorithm ends after it has explored the highest resolution (i.e. $D = 1$) (lines 6-12).

Note that when computing lower bounds at any resolution level, we take the pruned-off locations at the lower levels into account, meaning that we do not compute a lower bound for those regions. The lower bound computation process is described in TABLE II.

TABLE II. $LB_{KEOGH}DSMP$: COMPUTES THE $LB_{KEOGH}MP$ FOR THE DOWNSAMPLED TIME SERIES.

```
Algorithm: LB_KeoghDSMP
Inputs: time series T, subsequence length L, downsampling
factor D, the pruned locations of time series semantic
matrix index pruned
Outputs: the expanded LB_KeoghD:1MP and indexes LBMP, LB_index
1    T_D ← paa(T,D)
2    pruned_D ← paa(pruned,T_D)
3    L_D ← L×floor(length(T_D)/length(T))
4    MP_D, LB_index ← LB_Keogh(T_D,L_D,pruned_D)
5    LBMP ← interpolate(MP_D,floor(length(T)/length(T_D)))
6    LBMP ← sqrt(length(T)/length(T_D))× LBMP
7    end
```

Lines 1 and 2 downsample both the time series and the Boolean vector specifying the pruned and non-pruned locations. Line 3 scales downs the subsequence length relative to the downsampling rate. Lines 4-6 compute the downsampled lower bound $LB_{Keogh}D:1$, expand it to the size of the complete lower bound and scale up the result by the downsampling factor.

### D. Phase II

Let us review the situation at the end of phase I. From the original set of $n + L - 1$ candidate time series subsequences that *might* have contained the top-1 motif, we pruned many (hopefully the *vast* majority) of them into a much smaller set $c$, of remaining candidates.

Globally, we know:

1. A *best-so-far* value, which is an upper bound on the value of the top-1 motif. We also know which pair from $c$ is responsible for producing that low value.

Locally, for each subsequence, we know:

2. A DTW lower bound value on its distance to its nearest neighbor.

3. The location of its nearest neighbor in lower bound space, which may or may not also be its DTW nearest neighbor.

We now need to process the set of candidates $c$ to find the true top-1 motif, or it the current *best-so-far* refers to the top-1 motif, confirm that fact by pruning every other possible candidate.

Note that even if we processed all $O(c^2)$ pairwise comparisons randomly, there is still the possibility of pruning more candidates. In particular, every time the *best-so-far* value decreases, we can use the information in '3' above to prune additional candidates in $c$, whose lower bounds now exceed the newly decreased *best-so-far* value. As shown in TABLE III we can see this search as a classic nested loop, in which the outer loop considers each candidate in $c$, finding its DTW nearest neighbor (non-trivial match) in $c$.

Given our stated strategy of trying to drive the *best-so-far* down as fast as possible, the optimal ordering for our search is obvious. In the outer loop we should start with a candidate that is one of the true DTW motif pair, and the inner loop we should start with the other subsequence of that motif pair. Clearly, we cannot do this, since that assumes we already know what we are actually trying to compute. However, we can approximate this optimal ordering quite well. On average, the true DTW distance is highly correlated with its lower bound (see Fig. 8). Thus, we should order the outer loop in increasing order of the lower bounds provided by $LB_{Keogh}1:1$ in the last iteration of Phase I (line 2).

For the inner loop, for the very first iteration we consider the candidate's nearest neighbor in lower bound space and replace it with its immediate neighbor (4-7). After this first comparison, the subsequent iterations can be done in any order. The algorithm in TABLE III formalizes these observations.

TABLE III. SWAMP: DISCOVERS THE TOP-1 DTW MOTIFS.

```
Algorithm: SWAMP
Inputs: time series T, subsequence length L, warping window w
Outputs: top-1 DTW motifs distance and their locations
motif_pair, the candidate motif distance best-so-far
1    [LBMP,pruned,best-so-far,LB_index] ← ComputeDSMP(T,L,w)
2    [candids,candids_index] ← sorted(LBMP) // begin Phase II
3    for i=1:length(candids):
4      candid_idx ← candids_index[i]
5      if pruned[candid_idx]: continue
6      neigh_idx ← candid_idx+L:length(candids)
7      swap(neigh_idx[1], neigh_idx[LB_index[candid_idx]])
8      for j=1:length(neigh_idx):
9        if pruned[neigh_idx[j]]: continue
10       a ← T[candid_idx:candid_idx+L-1]
11       b ← T[neigh_idx[j]:neigh_idx[j]+L-1]
12       if LB_KimFL(a,b) >= best-so-far: continue
13       elseif LB_Keogh(a,b) >= best-so-far: continue
14       dist ← dtw_distance(a,b,w,best-so-far)
15       if dist < best-so-far:
16         best-so-far ← dist
17         motif_pair ← [candid_idx, neigh_idx]
18         pruned(candid_idx(candids >= best-so-far)) ← true
19   end
```

Note that we have added four further optimizations into the inner loop. We use a cheap but weak lower bound $LB_{Kim}FL$ to prune some subsequences (line 12). For those pairs that survive, we use a tighter but more expensive lower bound $LB_{Keogh}1:1$ (line 13). Moreover, we use the early abandoning version of $LB_{Keogh}$, as introduced in [8]. Finally, if all previous attempts at pruning fail, and we are forced to do DTW, we compute the early abandoning version of DTW, also introduced in [8] (line 14). If any candidates survive that step, we update the *best-so-far* and prune the remaining unpromising subsequences (line 15-18).

Revisiting $LB_{Keogh}1:1$ in this phase is worth clarifying. We do already know the $LB_{Keogh}1:1$ distance to each candidate's *nearest* neighbor (from Phase I), but not to *all* its neighbors. Therefore, it is possible that with another round of lower bound computation for the remaining pairs, we can potentially have more prunings.

## V. Experimental Evaluation

We begin by stating our experimental philosophy. We have designed all experiments such that they are easily reproducible. To this end, we have built a webpage that contains all datasets, code and random number seeds used in this work, together with spreadsheets which contain the raw numbers in addition to dozens of additional case studies and experiments [16]. This philosophy extends to all the examples in the previous section.

Note that in many works, the size of the warping window is given as a percentage of the length of the time series [3][9]; here we give it as an absolute number.

### A. Examples of DTW Motifs

Before conducting more formal experiments, we will take the time to show some examples of DTW motifs we have discovered in various datasets, in order to sharpen the readers' appreciation of the utility of DTW in motif discovery.

Entomologists use an apparatus called an electrical penetration graph (EPG) to study the behavior of sap-sucking insects [12]. It is known anecdotally [12], and by the use of classic motif discovery [6], that some such behaviors are often highly conserved at a time scale of 1 to 5 seconds. However, is there any behavior conserved at a longer time scale? As shown in Fig. 10.*bottom.left*, if we used the Euclidean distance, we might say "no". While the two patterns in the motif are vaguely similar, we might attribute this to random chance.

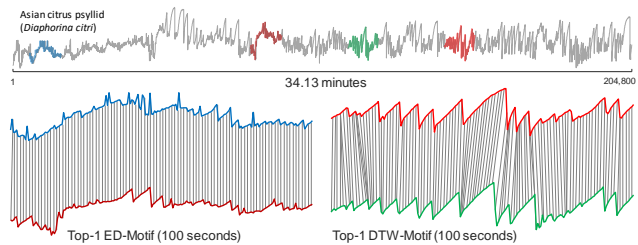

Fig. 10. *top*) About 34 minutes of EPG data collected from an Asian citrus psyllid (ACP) that was feeding on a Troyer citrange (Sweet Orange) [12]. *bottom*) The top-1 ED motif (*left*) and the top-1 DTW motif (*right*).

However, if we simply use the DTW distance, we discover an unexpectedly well-conserved long motif, corresponding to feeding behavior known as *phloem ingestion* [12]. Exploring such datasets rapidly gives one an appreciation as to how brittle the Euclidean distance can be. Consider the experiment on a different individual from the same species shown in Fig. 11.

As before, we cannot directly fault the ED. It *does* return a pair of subsequences that are similar, although somewhat "boring and degenerate". However, an entomologist would surely prefer to see the DTW motif, which contains examples of a probing behavior [12]. To understand why ED could not discover these, in Fig. 11.*bottom* we show the alignment both methods have on the sections corresponding to the behavior. ED, with its one-to-one alignment, cannot avoid mapping some peaks to valleys, incurring a large distance. In contrast, the flexibility of DTW allows it to map peak-to-peak and valley-to-valley, allowing the discovery of these semantically identical behaviors.

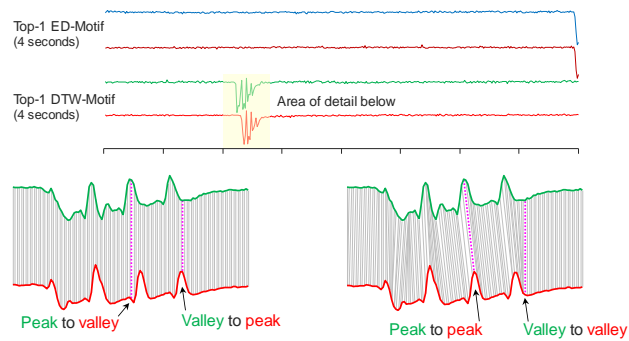

Fig. 11. *top*) The top-1 ED and DTW motifs discovered in seven-hour segment EPG data collected from an ACP [12]. *bottom*) A zoom-in of the DTW motif visually explains why ED has difficulty finding the same motif as DTW.

## VI. Conclusion and Future Work

We have introduced SWAMP, the first practical tool to find DTW-based motifs in large datasets, showing that on many real datasets, DTW returns more meaningful motifs.

## VII. Acknowledgment

This work is supported by gifts from MERL Labs, and a Google faculty award.